# COOR-PLT: A hierarchical control model for coordinating adaptive platoons of connected and autonomous vehicles at signal-free intersections based on deep reinforcement learning

Duowei LI[a,b], Feng ZHU[b], Jianping WU[a], Tianyi CHEN[b], Yiik Diew WONG[b]
a. Department of Civil Engineering, Tsinghua University, Beijing, China
b. School of Civil and Environmental Engineering, Nanyang Technological University, Singapore

**Abstract:** Platooning and coordination are two implementation strategies that are frequently proposed for traffic control of connected and autonomous vehicles (CAVs) at signal-free intersections instead of using conventional traffic signals. However, few studies have attempted to integrate both strategies to better facilitate the CAV control at signal-free intersections. To this end, this study proposes a hierarchical control model, named COOR-PLT, to coordinate adaptive CAV platoons at a signal-free intersection based on deep reinforcement learning (DRL). COOR-PLT has a two-layer framework. The first layer uses a centralized control strategy to form adaptive platoons. The optimal size of each platoon is determined by considering multiple objectives (i.e., efficiency, fairness and energy saving). The second layer employs a decentralized control strategy to coordinate multiple platoons passing through the intersection. Each platoon is labeled with coordinated status or independent status, upon which its passing priority is determined. As an efficient DRL algorithm, Deep Q-network (DQN) is adopted to determine platoon sizes and passing priorities respectively in the two layers. The model is validated and examined on the simulator *Simulation of Urban Mobility* (SUMO). The simulation results demonstrate that the model is able to: (1) achieve satisfactory convergence performances; (2) adaptively determine platoon size in response to varying traffic conditions; and (3) completely avoid deadlocks at the intersection. By comparison with other control methods, the model manifests its superiority of adopting adaptive platooning and DRL-based coordination strategies. Also, the model outperforms several state-of-the-art methods on reducing travel time and fuel consumption in different traffic conditions.

## Symbols and abbreviations

| | |
|---|---|
| CAV | Connected and autonomous vehicle |
| HV | Human-driven vehicle |
| AIM | Autonomous intersection management |
| DRL | Deep reinforcement learning |
| IM | Intersection manager |
| LVA | Leader vehicle agent |
| FVA | Follower vehicle agent |
| DQN | Deep Q-Network |
| $G$ | DRL agent |
| $W$ | Environment |
| $s$ | State |
| $a$ | Action |
| $r$ | Reward |
| $Q$ | Action-value function |
| $\tilde{P}_i$ | The $i$-th platoon |
| $\mathcal{S}_1$ | State space in the first layer |
| $\mathcal{A}_1$ | Action space in the first layer |
| $\mathcal{R}_1$ | Reward space in the first layer |
| $d_h$ | Desired distance headway in the platoon |
| $l_c^i$ | Length of the $i$-th CAV |
| $W_i$ | Waiting time of the $i$-th CAV |
| $PW_i$ | Penalized waiting time of the $i$-th CAV |
| $D_i$ | Delay of the $i$-th CAV |
| $F_i$ | Fuel consumption of the $i$-th CAV |
| $\mathcal{S}_2$ | State space in the second layer |
| $\mathcal{A}_2$ | Action space in the second layer |
| $\mathcal{R}_2$ | Reward space in the second layer |
| $g$ | Granularity |
| $CT$ | Coordination time |
| $PC_i$ | Travel time of the $i$-th CAV in coordination zone after coordination |
| $R_{deadlock}$ | Punishment of deadlock |

# 1 Introduction

With the improvement of people's living standards, higher transport demand has brought enormous challenges to urban transportation system. An ever-increasing number of conventional human-driven vehicles (HVs) would escalate further on serious traffic problems such as crash risk, traffic delay, and energy consumption. The impacts of those problems are amplified as traffic conditions at the intersections are more complex (Namazi et al., 2019). Hence, for the purpose of safety enhancement and operational efficiency, intersection management has drawn much attention from academia and industry. Recently, the emergence of connected and autonomous vehicles (CAVs) has provided fitting opportunity to address the above traffic problems (Zhang et al., 2021). Thus, how to better involve CAVs into intersection management has become a focus of researchers.

There are two main directions of involving CAVs to improve intersection management in previous studies. On one hand, some studies improved conventional traffic signal control strategy by dynamically adjusting the signal lights in terms of phase cycle and sequence, using the real-time spatial-temporal traffic data provided by CAVs (Li et al., 2020). Those strategies have adopted conventional signalized intersection management which is suitable for traffic condition with a low penetration rate of CAVs. However, the signal control at intersections is not necessarily the optimal strategy for a pure CAV environment (where all the vehicles are CAVs), since traffic signal may cause unnecessary waiting time (Chai and Wong, 2015). On the other hand, some researchers have conducted studies on an intersection management concept customized for signal-free intersections, namely, Autonomous Intersection

Management (AIM) (Dresner and Stone, 2008; Meng et al., 2017). As compared to HVs, CAVs have several advantages, such as higher predictability of movement, shorter distance headway, and quicker responses to emergencies, which can further ensure both traffic safety and efficiency at signal-free intersections (Xu et al., 2021). Hence, AIM is a strategic concept that is suitable for pure CAV environment at signal-free intersections.

Platooning and coordination are two concrete implementation strategies that can be adopted by AIM. Platooning is a method for grouping vehicles with similar attributes into an indivisible queue (Cao et al., 2021), within which the vehicles move in the same direction while maintaining a stable distance headway (Berbar et al., 2022; Zhou and Zhu, 2021). The control of CAVs at signal-free intersections can be categorized into individual CAV control (Buzachis et al., 2020; Čakija et al., 2019; Dresner and Stone, 2008; Wu et al., 2019) and CAV platoon control (Bashiri et al., 2018; Kumaravel et al., 2021; Li et al., 2020). As compared to individual CAV control, CAV platoon control at signal-free intersection is more complicated as it encapsulates both centralized control and decentralized control, but it can provide more benefits to AIM due to its following characteristics: (1) reduced communication burden: the intersection manager (IM) only communicates with the platoon leader instead of all individual CAVs (Kumaravel et al., 2021); (2) higher operational efficiency: close and insert-forbidden vehicle-following behavior is required on each CAV (Lioris et al., 2017); and (3) lower energy consumption: platooning reduces the frequency of CAVs' acceleration and deceleration (Lesch et al., 2021).

Coordination indicates a strategy that multiple CAVs jointly decide their movements or trajectories at intersections via V2V communication (Thormann et al., 2022). Some previous studies failed to take coordination into consideration (Bashiri et al., 2018; Dresner and Stone, 2008; Jin et al., 2013; Kumaravel et al., 2021). Those studies adopted the CAV releasing rule that merely allows nonconflicting movement at intersections, which scarifies traffic efficiency over safety. Although CAV coordination is demanding due to its complicated CAV releasing rules and collision avoidance schemes, several previous studies (Chen et al., 2022; Wu et al., 2019; Xu et al., 2018) have demonstrated its superiority in solving AIM problems. The contributions of CAV coordination can be explained from the following two aspects: (1) throughput of intersection area can be improved as those CAVs having conflicting movements are allowed to simultaneously enter the intersection; and (2) travel delays can be reduced since unnecessary waiting before the stop line is avoided.

Herein, involving platooning and coordination into AIM are challenging but worthy of detailed investigation (Thormann et al., 2022). Although much effort has been made to verify the reliability and effectiveness of platooning and coordination in previous studies, there are still several limitations in those studies: (1) few studies have attempted to encapsulate coordination into CAV platoon control; (2) most existing models on CAV platoon control adopt inefficient platoon releasing rules; and (3) coordination models proposed in previous studies are applicable to limited traffic conditions. The above-mentioned limitations are discussed in detail in Section 2. To resolve those limitations, this study proposes a deep reinforcement learning (DRL) powered hierarchical control model for coordinating adaptive CAV platoons at signal-free intersections, named COOR-PLT. In this model, DRL technique is adopted to handle the two control tasks, namely, adaptive platoon formation and platoon coordination. DRL techniques have made promising progress for conventional traffic signal control, as they are able to address complicated tasks with less prior knowledge. However, they have not been applied in the AIM problem that involves coordination of CAV platoons. The proposed model in this study has the following innovations:

(1) A two-layer hierarchical model structure, which comprises adaptive platoon formation layer and platoon coordination layer, is adopted to realize coordination of CAV platoons at signal-free intersections. The two layers respectively implement centralized control and decentralized control strategy. The proposed model can be deployed in more complex traffic conditions under the premise of being collision-free.

(2) An efficient platoon releasing rule, which adopts adaptive platooning and allows platoons

from all lanes to enter the coordination zone simultaneously, is proposed to increase the throughput of the intersection. A DRL algorithm is applied to determine the optimal size of each platoon while considering multiple objectives (i.e., efficiency, fairness and energy saving).

(3) The coordination of CAV platoons is modelled as a multi-agent DRL problem, in which each platoon is controlled by its leader as an agent. The issues related to the coordination, such as determination of granularity and avoidance of deadlock, are well addressed. The platoons are labeled with coordinated status or independent status, the passing priorities of which are determined by a DRL algorithm to reduce computational complexity.

The rest of this paper is organized as follows. Section 2 reviews related work. Problem statement is presented in Section 3. In Section 4, the model framework including the two layers is introduced. Section 5 validates the proposed model on a simulator, and Section 6 covers the conclusion of this study.

# 2 Literature review

Involving CAVs into the control of signal-free intersections can sufficiently enhance both safety and efficiency of intersection operations. However, controlling intersections via traffic signals as a conventional method would no longer be an optimal strategy under pure CAV environment (Wu et al., 2021). Thus, many research studies have been dedicated in developing advanced strategies for pure CAV environment at intersections. Dresner and Stone (2004) first proposed a protocol controlling CAVs at signal-free intersections, i.e., Autonomous Intersection Management (AIM), which plays a role as a cornerstone for the research relevant to signal-free intersection control involving CAVs. After then, several attempts have been made to address the problems in the main branches of AIM, namely, centralized and decentralized control strategies.

In centralized control strategy, a central controller (i.e., intersection manager (IM)) is set in the intersection to make at least one decision for all CAVs (Rios-Torres and Malikopoulos, 2017). There are two commonly-used approaches to implementing centralized control strategies, namely, reservation-based methods and objective optimization. The AIM protocol proposed in Dresner and Stone (2004) and its further modified version (Dresner and Stone, 2008) are reservation-based control methods. The protocol controls CAVs to simultaneously pass through an intersection by applying a First Come First Served (FCFS) policy to deal with requests by CAVs. To improve the performance of reservation-based AIM, many researchers (Buzachis et al., 2020; Huang et al., 2012; Zhang et al., 2015) have refined the protocol in different aspects. However, the performances of most reservation-based methods are limited as an IM can only process one request at a time and the communication burden is always heavy. Other studies adopt objective optimization to conduct centralized control strategies. For example, Wu et al. (2014) used dynamic programming method to identify the passing priorities that minimize the travel time given the speed and position of CAVs. Lee and Park (2012) developed an optimized model with the objective to minimize the overlap of vehicle trajectories, which reduces delays and $CO_2$ emissions in comparison with reservation-based model. However, although the IM in those methods that are purely based on centralized control strategies can make a global-optimal decision for the CAVs at an intersection, comprehensive information from the intersection is always required in the method, which leads to high computational and storage expense (Xu et al., 2018). Therefore, merely adopting centralized control strategy is not an appropriate way for CAV platooning and coordination at intersections.

In decentralized control strategy, a central controller is no longer needed. Instead, CAVs plan their trajectories in a coordinated manner through vehicle-to-vehicle (V2V) communication. The most distinctive difference between centralized and decentralized control strategies is that the former strategy relies on global information of all the CAVs at an intersection, while the latter one applies local information within the detection and communication range of each individual CAV (Zhou et al., 2022). As compared to centralized control, although decentralized control may not attain global optimization, but it can sufficiently improve computational efficiency as well as reduce the deployment and maintenance costs of infrastructure. Thus, more researchers have adopted decentralized strategies in solving AIM problems under relative complex traffic conditions. Also, objective optimization is a technique that has been widely used in previous research on decentralized control. The objectives include avoiding collisions, minimizing speed deviations, minimizing safe headway spacing, and maximizing speed and acceleration (Kim and Kumar, 2014; Mahbub et al., 2020; Makarem and Gillet, 2013; Yao and Li, 2020). Recently, several studies have also attempted to use reinforcement learning (RL) technique to address the problems related to decentralized AIM. As compared to objective optimization, RL technique is more adept at adaptive control tasks in dynamic traffic conditions. Mokhtari and Wagner (2021) proposed a deep reinforcement learning (DRL) approach for individual CAV to make a left turn without crashing into pedestrian at signal-free intersections. Similarly, Qiao et al. (2018) adopted a DRL model to control a CAV to safely bypass another vehicle at a stop-sign intersection. However, the assumptions made in the above two studies are rather simple, which only involves the control of an individual CAV without considering coordination. To resolve this limitation, Wu et al. (2019) proposed a RL-based decentralized CAV coordination model for signal-free intersection, i.e., DCL-AIM. It regards the movements of the CAVs passing through a signal-free intersection as multi-agent Markov decision processes, in which CAVs adopt RL technique to adjust their acceleration and to minimize intersection delay under the constraints of being collision-free. Nevertheless, the DCL-AIM model can only deal with the coordination between two individual CAVs, as the Q-learning algorithm employed by the model is not suitable for complex traffic conditions.

To synthesize the advantages f centralized and decentralized control strategies, several hierarchical models combining those two strategies have been proposed so far. In those hierarchical models, the models with two-level framework that involve CAV platoons have been widely used to achieve global optimization while reducing computational complexity. For example, Jin et al. (2013) proposed a two-level model where reservation process is conducted between IM and leader vehicle agents (LVAs) at the first level and LVA plans the trajectory for follower vehicle agents (FVAs) based on the reservation information at the second level. Bashiri and Fleming (2017) developed a similar two-level model that can derive optimal schedules for platoons with minimal traffic delay. Kumaravel et al. (2021) proposed a hierarchical two-level model that allows each LVA to communicate with others, upon which the central coordinator can derive an optimal schedule for platoons to pass through the intersection with least waiting time. However, those existing two-level models are not flexible and efficient as they adopt fixed length platoon and do not consider coordination among platoons.

In effect, previous studies have made various contributions to CAV control at signal-free intersections by adopting either platooning or coordination strategies. However, few studies have ever attempted to integrate both strategies into CAV control at the intersections. Further, there are still several research gaps that remain to be addressed, which are summarized as follows:

- In those studies that encapsulate platooning into signal-free intersection control, there are two gaps that may greatly limit traffic efficiency at intersections. First, the adopted platoon releasing rules are rather conservative, as some studies allowed only one platoon to pass through the intersection at a time, while the others merely considered nonconflicting movement when releasing platoons. Second, most of those studies adopted fixed length platoon without comprehensively considering the relations between CAV platoon size and traffic conditions around an intersection.

- The main gap in the studies that consider CAV coordination is that the proposed coordination strategies are applicable to a limited range of traffic conditions. For example, some strategies adopt the predefined policy that fails to handle dynamic traffic conditions, while the other strategies based on non-predefined policy can only realize the coordination between a limited number of CAVs.

To the best of our knowledge, this study is one of the first attempts to propose a model that adopts both adaptive platooning and coordination strategies to control CAVs at signal-free intersections. Moreover, the above-mentioned research gaps are resolved in this study for a better implementation of the proposed model.

# 3 Problem statement

## 3.1 Problem description

In this study, we only consider pure CAV environment at a signal-free intersection, where CAV platoons are coordinated to pass through the intersection. As illustrated in Figure 1, to simplify the description, this study separates an intersection area into two zones, namely formation zone and coordination zone. In the formation zone, the CAV platoon is composed of a leader vehicle agent (LVA) and several follower vehicle agents (FVAs) and shall be formed before the stop line. The coordination zone is a square with length equal to $S$, in which the coordination among CAV platoons occurs. An intersection manager (IM) is deployed at the center of the intersection. It is worth noting that the coordination zone is not included in the formation zone. For a clearer representation, the individual CAVs that have not joined a platoon are marked with grey color, while the CAVs belonging to different platoons are distinguished by being marked with different bright colors.

There are two critical problems to be solved in this study. **The first problem** is how to determine the optimal platoon size for CAVs and form the platoon in each lane of the formation zone based on the global real-time traffic information (e.g., traffic density, vehicles' movement, etc.) collected by IM. **The second problem** is how to coordinate multiple CAV platoons to pass through the coordination zone simultaneously without collision based on the local information (e.g., platoon position, speed, etc.) shared between the platoons. The two problems shall be addressed with the objective to make a trade-off between safety, traffic efficiency and energy saving at the intersection area.

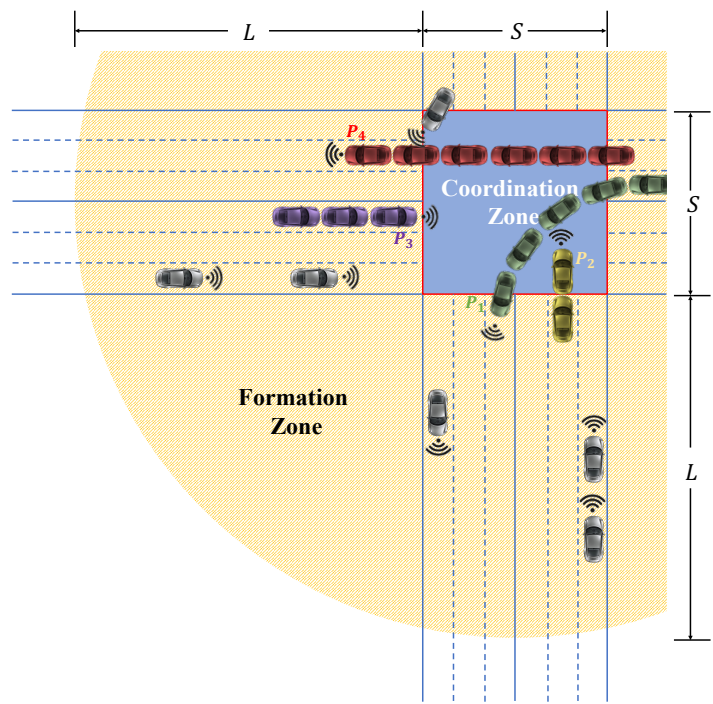

Figure 1. Illustration of intersection area.

## 3.2 Basic definitions and assumptions

In this study, the following basic definitions are made to facilitate the solutions of above-mentioned problems.

**Definition 1: Conflict movements.** The platoons entering the coordination zone along their pre-defined trajectories may have conflicts with others. We define the movements of any two platoons as conflict movements if their pre-defined trajectories intersect. Herein, the point where two movements intersect is defined as a **conflict point (CP)**. Taking the South-North movement (green colored) illustrated in Figure 2(a) as an example, its four conflict movements as well as the corresponding CPs are respectively presented in Figures 2(b)-2(e).

**Definition 2: Conflict platoons.** The platoons having conflict movements are defined as conflict platoons. For example, as shown in Figure 1, the platoon $P_1$ in green and the platoon $P_2$ in yellow are two conflict platoons. The conflict between the platoons $P_1$ and $P_2$ can be described as the scenario in Figure 2(c).

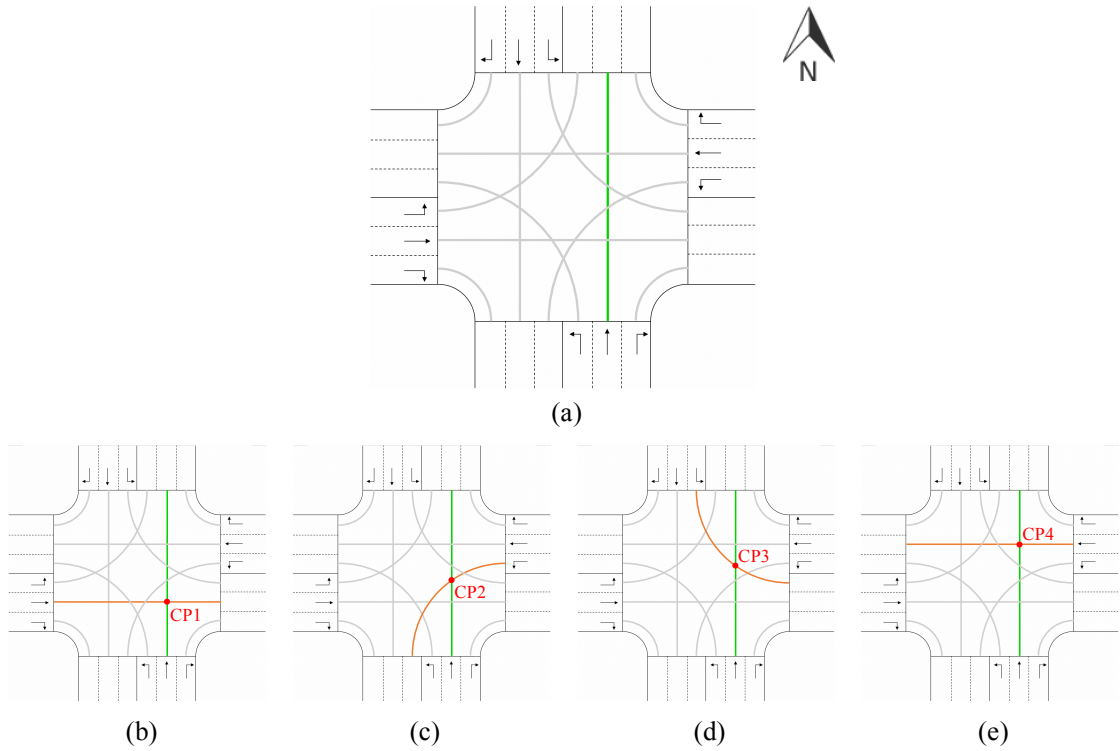

Figure 2. Illustrations of conflict movements.

**Definition 3: Minimum coordination distance (MCD).** Conflict platoons shall be coordinated to avoid collisions in the system. The coordination decision should be made early enough so that the platoons can have sufficient distance to decelerate and stop before reaching the CP. Herein, we define minimum stopping distance (MSD) as the distance with which a platoon travels at the maximum deceleration rate till it stops at the CP. The MCD of a pair of conflict platoons can be obtained by Equation (1), where $d_1$ and $d_2$ respectively denote the MSDs of the two platoons.

$$MCD = min\,(d_1, d_2) \qquad (1)$$

**Definition 4: Granularity.** To ensure that the environment is precisely represented, the coordination zone is divided into $g \times g$ grids, where $g$ is defined as the granularity of the model (Dresner and Stone, 2004). Herein, the traffic environment (e.g., platoon's location, speed, etc.) can be represented based on the grids that the platoons are occupying. Figure 3 illustrates two examples of gridding the coordination zone with different granularities (i.e., 6 and 12), where the two corresponding matrixes respectively present the locations of the two platoons.

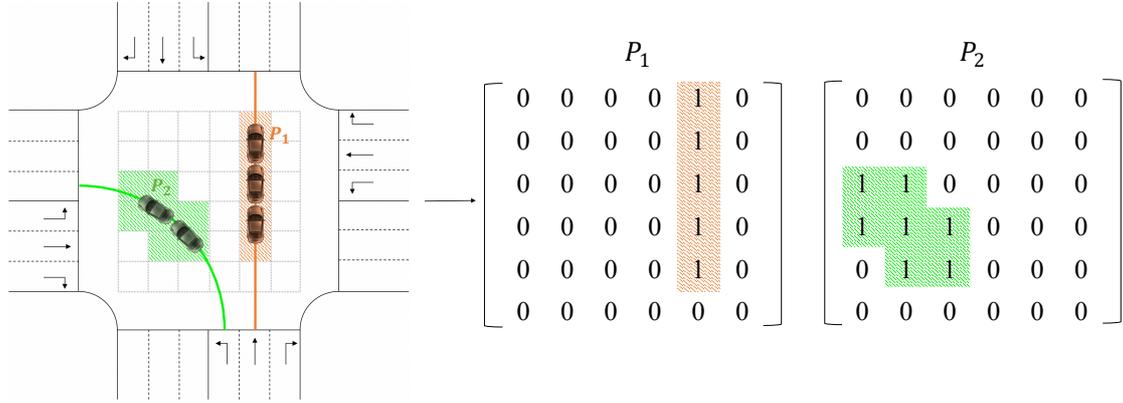

(a) Granularity = 6

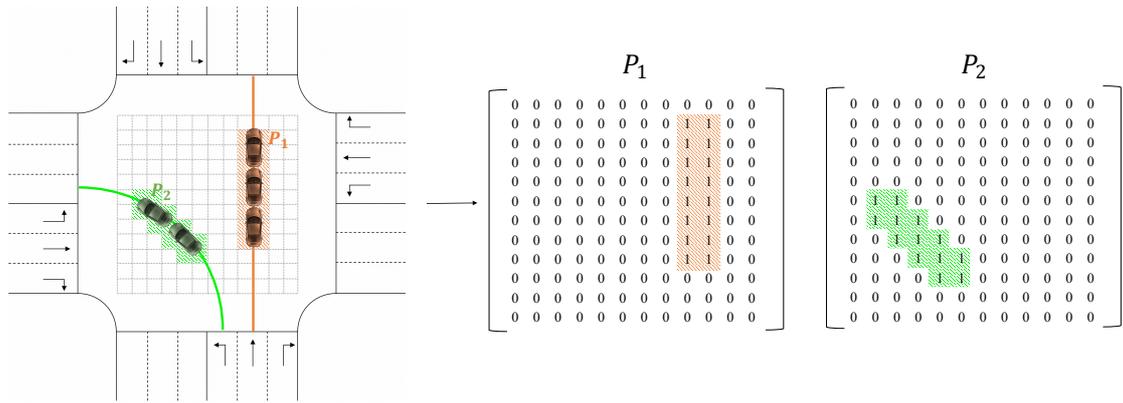

(b) Granularity = 12

Figure 3. Illustrations of gridding the coordination zone with different granularities.

Additionally, the following specific assumptions are imposed to solve the problems:

(1) Communication delay and errors between IM, LVAs and FVAs are not considered.

(2) CAVs within a platoon shall travel with the same speed and maintain a constant desired distance headway.

(3) Platoons should travel along a pre-defined trajectory in the coordination zone as shown in Figure 2.

(4) Lane-changing maneuvers are not allowed in the formation zone, which means all CAVs should proactively move to a desired lane before entering the formation zone. Also, the CAVs shall send a request to the IM when entering the formation zone.

(5) In each moving direction, only one platoon is permitted to travel in the coordination zone. In other words, in each lane, the following platoon shall wait before the stop-line until its preceding platoon exits the coordination zone.

# 4 Methodology

## 4.1 Hierarchical framework

To address the above mentioned two problems, we develop a hierarchical model namely, COOR-PLT, which integrates platooning and coordination into CAV control at a signal-free intersection. The proposed model has a framework composed of two layers as represented in Figure 4.

The first layer refers to adaptive platoon formation layer, which mainly addresses the first problem by using centralized control strategy. There are three main steps (i.e., Steps 1 to 3) in this layer. In Step 1, when a CAV approaches the formation zone, it transmits a request to IM, which contains information such as vehicle's ID, position, speed, and acceleration / deceleration limits. In Step 2, the optimal platoon size for each lane is determined by the IM (controlled by a DRL agent). In Step 3, the IM directs CAVs in each lane to form platoons with each platoon having its optimal size.

The second layer indicates platoon coordination layer, in which the second problem is solved based on decentralized control strategy. There are two steps (i.e., Steps 4 and 5) in this layer. In Step 4, the status of each platoon in the coordination zone is scanned, identified, and tracked all the time. There are two platoon statuses, namely, coordinated status and independent status. The conflict platoons are labeled with "coordinated status", and their coordination actions are determined jointly and performed separately. Since each platoon is controlled by an independent DRL agent, the coordination process can be regarded as decentralized multi-agent coordination. Apart from the conflict platoons, the other platoons without conflict movements are labeled with "independent status" and therein perform independent action. Additionally, for each platoon, the two steps in this layer are executed in a loop until the platoon exits the coordination zone.

In this study, a communication protocol plays the cornerstone role for modeling the COOR-PLT, as several critical steps adopt the information transmission criteria of the protocol. As listed in Table 1, the communication protocol contains the messages transmitted between the LVAs, FVAs, and IM. As shown in Figure 4, the four categories of transmissions, namely, CAV to IM, IM to LVA, LVA to FVA, and LVA to LVA, occur sequentially once a CAV enters the formation zone. The two layers will be respectively introduced in Sections 4.3 and 4.4 in details.

Table 1. Message categories in communication protocol.

| Message categories | Content |
| --- | --- |
| CAV to IM | Vehicle ID; Current position; Current speed; Acceleration/ deceleration limit; Turning demand |
| IM to LVA | Platoon size; Desired speed; Desired head spacing |
| LVA to FVA | Join or not; Head spacing; Follow speed |
| LVA to LVA | Vehicle ID; Current position; Current speed; Acceleration/ deceleration limit; Turning demand |

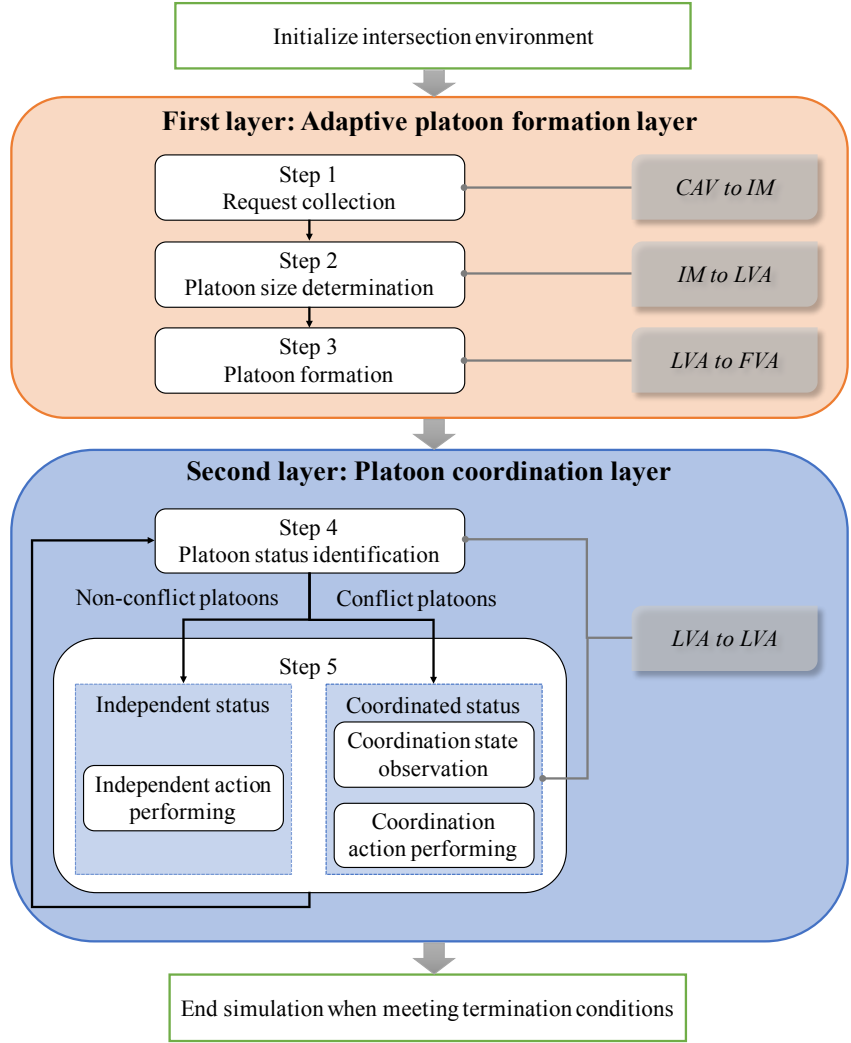

Figure 4. Hierarchical framework of COOR-PLT.

### 4.2 DRL techniques

In this study, deep reinforcement learning (DRL) techniques are used to implement the two layers, which play a role as a backbone of the proposed model. Reinforcement learning (RL) is a category of machine learning techniques that determines how agents take actions to maximize expected benefits in an environment (Sutton and Barto, 2018). As shown in Figure 5, an RL procedure is mainly concerned with the interaction between the agent $G$ and the environment $W$, which works by executing a loop with following three steps: (1) a state $s$ is retrieved by $G$ from $W$; (2) an action $a$ is selected by $G$ based on the current policy and is performed on $W$; and (3) a reward $r$ is calculated by $W$ and transmitted to $G$ for updating the policy. The loop is ended when a terminal condition is met. This procedure is regarded as a Markov Decision Process (MDP), which can be expressed as:

$$\mathcal{M} \stackrel{\text{def}}{=} \langle \mathcal{S}, \mathcal{A}, \mathcal{P}, \mathcal{R} \rangle \tag{2}$$

where state space $\mathcal{S}$, action space $\mathcal{A}$, and reward space $\mathcal{R}$ respectively indicate the finite sets of $s$, $a$, and $r$, and $\mathcal{P}$ is the state transition policy (Li et al., 2022). RL technique is used to identify the optimal policy $\mathcal{P}^*$, with which the maximum of $r$ is obtained, by learning the action-value function $Q$. The action-value function at the timestamp $t+1$ can be expressed as:

$$Q_{t+1}(s,a) = Q_t(s,a) + \alpha \left[ r + \gamma \max_{a'} Q_t(s',a') - Q_t(s,a) \right] \quad (3)$$

where $\alpha$ ($0 \leq \alpha \leq 1$) refers to learning rate, which determines the proportion of the existing information that is updated by the newly acquired, and $\gamma$ ($0 \leq \gamma \leq 1$) refers discount factor, which determines the importance of future rewards. Accordingly, the optimal policy $\mathcal{P}^*$ shall satisfy the following equation:

$$Q^*(s,a) = \max_{\mathcal{P}^*} Q(s,a) \quad (4)$$

where $Q^*$ indicates the action-value function given the optimal policy $\mathcal{P}^*$.

Trial-and-error search and delayed reward are the two important functional properties that enable RL to be competent for handling the two adaptive control tasks in our proposed model. Figure 5 also illustrates the basic settings of RL in the two layers of the proposed model. In the first layer, the items $G$, $a$, and $W$ respectively represent IM, (identifying) platoon size, and traffic condition in the formation zone, while in the second layer, the three items respectively indicate LVA of platoon, (determining) passing priorities, and traffic condition in the coordination zone. Herein, the items $s$, $a$, and $r$, which respectively denote state, action, and reward, are the three most essential components of an RL agent. In each layer, the three finite sets of them, namely, $\mathcal{S}$, $\mathcal{A}$ and $\mathcal{R}$ are defined according to the layer function.

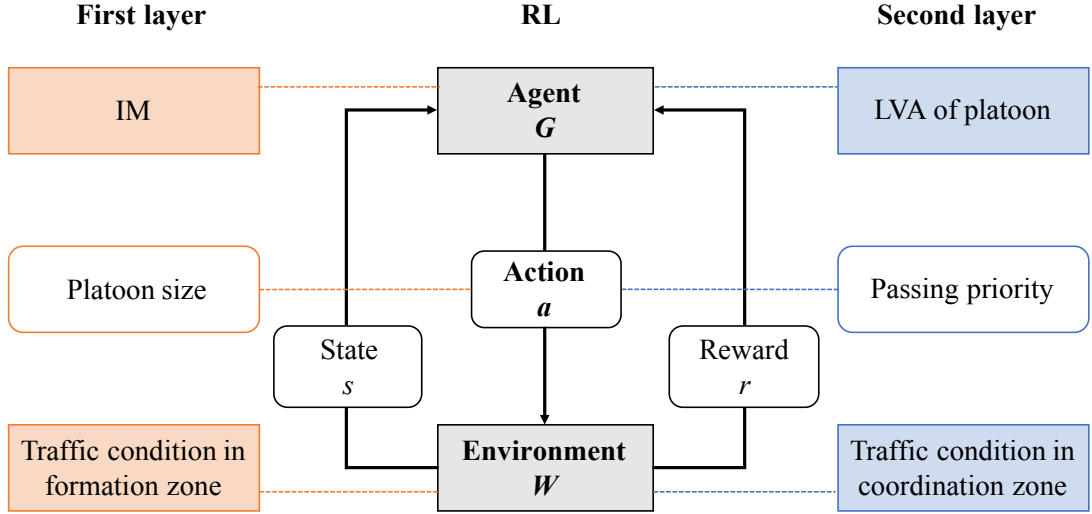

Figure 5. Procedure of RL and its applications in COOR-PLT.

As an improved RL technique, DRL encapsulates deep neural network as an action-value function $Q$ approximator. DRL has better performances when being trained in a complex and large-scale environment as compared to conventional RL techniques. In this study, as an DRL technique, Deep Q-Network (DQN) (Mnih et al., 2015) is used to determine the optimal platoon size and passing priorities for platoons. Herein, convolutional neural network (CNN) (Krizhevsky et al., 2012) is employed as the action-value function approximator embedded in DQN. CNN is able to effectively capture the traffic characteristics of an intersection due to its capability of being sensitive to spatial features. As illustrated in Figure 6, in this study, the CNN capsules input state matrixes, multiple convolutional layers, optional hidden layers (i.e., fully connected layers), and an output action-value list. Herein, the features that contribute to the output are extracted in the convolutional layers (Yu et al., 2020). The transforming process of the convolutional layers is defined as (O'Shea and Nash, 2015):

$$y_{i^{l+1},j^{l+1},d}^{l+1} = \sum_{i=0}^{h^l}\sum_{j=0}^{w^l}\sum_{d^l=0}^{D^l} f_{i,j,d^l,d} \times x_{i^{l+1}+i,j^{l+1}+j,d^l}^{l} \qquad (5)$$

where $x_{i^{l+1}+i,j^{l+1}+j,d^l}^{l}$ is the source pixel of the $l^{th}$ layer, and $y_{i^{l+1},j^{l+1},d}^{l+1}$ is the destination pixel of the $(l+1)^{th}$ layer. For example, if the input of the $l^{th}$ convolutional layer is an third-order tensor with the size $H^l \times W^l \times D^l$ and the size of kernel tensor $f$ is $h^l \times w^l \times D^l \times d^l$, the size of the output tensor will be $(H^l - h^l + 1) \times (W^l - w^l + 1) \times d^l$.

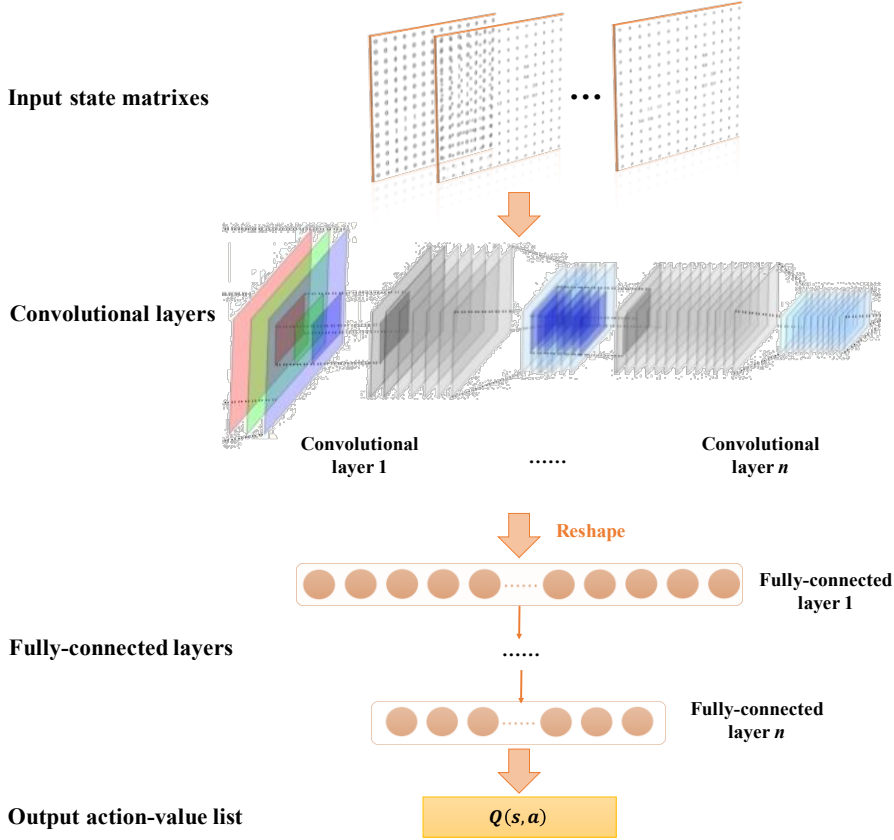

Figure 6. Structure of embedded CNN in DQN.

The general procedure of applying DQN is presented in Algorithm 1. The DQNs are respectively implemented for the two layers based on the general procedure. The actions are selected by adopting an ε-greedy strategy, which assumes that the probability of random exploration of the environment is $\varepsilon$, while the probability of selecting the action with highest value equals 1-$\varepsilon$. Herein, the network parameter set $\theta$ is updated by performing a stochastic gradient descent step on the loss function, which is defined as $L = (y_j - Q(s_j, a_j; \theta))^2$ and minimized by Adam algorithm (Zhang, 2018). Also, the following two modifications are made to make DQN suitable for training large-scale CNNs without diverging: (1) using experience replay; and (2) adopting two separate networks, namely, Target Net and Evaluation Net. More detailed explanations can be found in our previous work (Li et al., 2022). It is worth noting that, in this study, the detailed procedure of DQN and the structure of the embedded CNNs are designed in accordance with the specific problem settings in each layer, which are discussed in the following sections.

| | |
|---|---|
| **Algorithm 1** General DQN procedure in COOR-PLT | |
| 1. | **Definition** |
| 2. | $D :=$ replay memory pool |
| 3. | $N :=$ maximum number of experiences in $D$ |
| 4. | $Q :=$ action-value function in Evaluation Net |
| 5. | $\hat{Q} :=$ action-value function in Target Net |
| 6. | $M :=$ maximum number of episodes |
| 7. | $T :=$ maximum number of iterations in each episode |
| 8. | **Initialization** |
| 9. | $D \leftarrow$ Initial replay memory to capacity $N$ |
| 10. | $Q \leftarrow$ Initial evaluate action-value function with random weights $\theta$ |
| 11. | $\hat{Q} \leftarrow$ Initial target action-value function with random weights $\theta^- = \theta$ |
| 12. | **For** episode $= 1, M$ **do** |
| 13. | Observe $O$ steps before decision-making |
| 14. | Initialize environment state $s_1$ |
| 15. | **For** $t = 1, T$ **do** |
| 16. | With probability $\varepsilon$ select a random action $a_t$ |
| 17. | Otherwise select $a_t = argmax_a Q(s_t, a; \theta)$ |
| 18. | Execute action $a_t$ and observe reward $r_t$ and environment state $s_{t+1}$ |
| 19. | Store experience $e_t = (s_t, a_t, r_t, s_{t+1})$ in $D$ |
| 20. | Sample random $batch\_size$ experiences $e_j = (s_j, a_j, r_j, s_{j+1})$ from $D$ |
| 21. | Set $y_j = \begin{cases} r_j, \text{ if episode terminates at step j+1} \\ r_j + \gamma max_{a'} \hat{Q}(s_{j+1}, a'; \theta^-), \text{ otherwise} \end{cases}$ |
| 22. | Updating network parameters $\theta$ by a gradient decent step on $(y_j - Q(s_j, a_j; \theta))^2$ |
| 23. | Reset $\hat{Q} = Q$ in every $C$ steps |
| 24. | Set $s_t = s_{t+1}$ |
| 25. | **End for** |
| 26. | **End for** |

## 4.3 First layer: Adaptive platoon formation layer

This layer adopts a modified version of the method proposed in our previous work (Li et al., 2022) to determine the optimal platoon size in each lane. In the previous work, a reservation-based strategy was employed for developing the CAV releasing rule. In that strategy, the platoon in the target lane, which is identified based on a FCFS policy, and the CAVs in the nonconflicting lane are allowed to pass through the intersection simultaneously. However, as mentioned in Section 2, this platoon releasing rule is less efficient, since coordination between platoons is not considered in this rule, which might cause the space of the intersection not being sufficiently utilized. In addition, the method in the previous work only considers a single and fixed optimization objective (i.e., minimize waiting time). Herein, to take platoon coordination as well as multiple objectives (i.e., efficiency, fairness and energy saving) into account, this study proposes an improved method for this layer based on our previous work, which is discussed in the following sub-section.

### 4.3.1 General process

In this study, the adaptive platoons in all the lanes of an intersection shall be formed in the formation zone, and they are allowed to enter the coordination zone simultaneously. As presented in Figure 4, this layer comprises three steps, namely, request collection, platoon size determination, and platoon formation. The agent (i.e., IM) of this layer is responsible for monitoring the environment, determining the optimal platoon size, and guiding CAVs to form platoon in each lane. To simplify the description, we select a lane being controlled by the IM as target lane. As illustrated in Figure 7, each target lane is processed by following the three steps. In this layer, platoon size determination by adopting DQN is the most pivotal step. The agent generates four input matrixes using the information in the requests sent from CAVs (as mentioned in Table 1). After processing the input matrixes through a CNN, the agent outputs an optimal platoon size. Once the platoon with the optimal size is formed and passes through the coordination zone, the next round of the platoon formation process will start. The parameters of the CNN used in this layer are presented in Table 2.

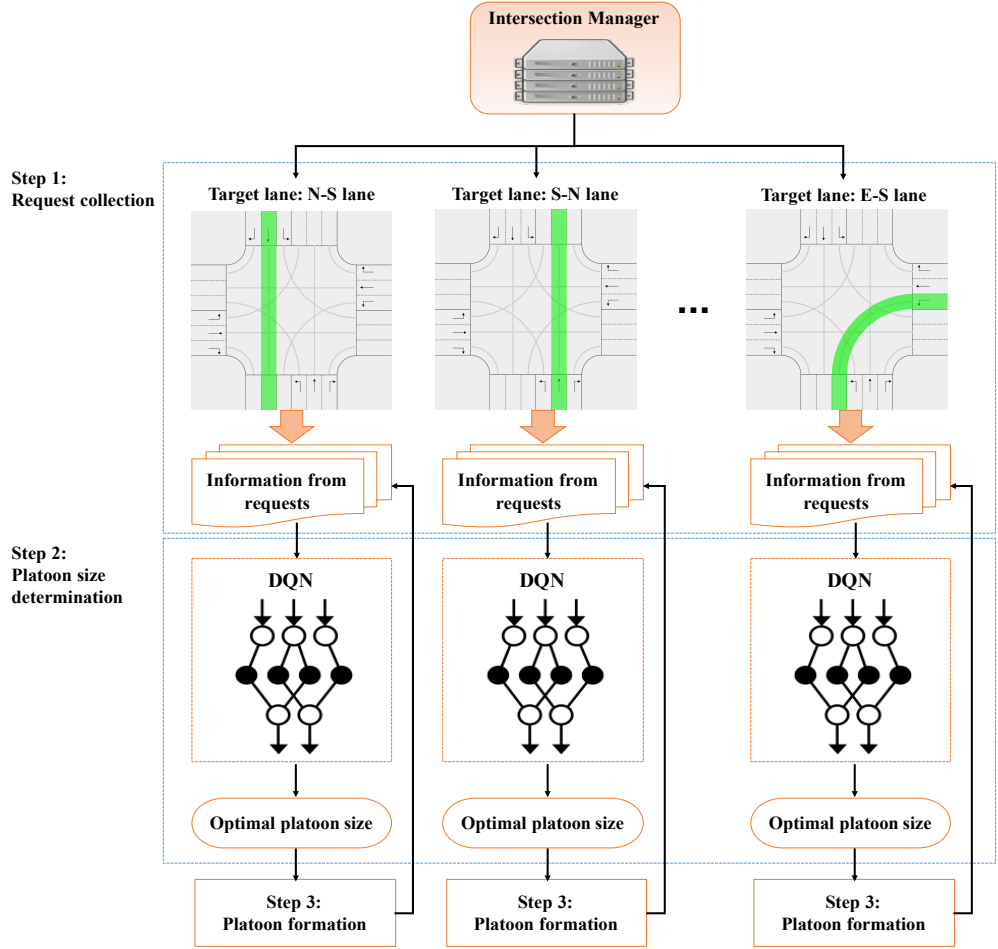

Figure 7. Flowchart of adaptive platoon formation process.

Table 2. Parameters of CNN in the first layer.

| DQN layers | Input size | Kernel size | Strides | Filters | Output size |
| --- | --- | --- | --- | --- | --- |
| Convolution 1 | 160×160×4 | 5×5 | 3×3 | 32 | 52×52×32 |
| Max Pooling | 52×52×32 | - | 2×2 | - | 26×26×32 |
| Convolution 2 | 26×26×32 | 3×3 | 2×2 | 32 | 12×12×32 |
| Padding | 12×12×32 | - | - | - | 26×26×32 |
| Convolution 3 | 26×26×32 | 3×3 | 2×2 | 64 | 12×12×64 |
| Convolution 4 | 12×12×64 | 3×3 | 2×2 | 64 | 5×5×64 |
| Reshape | 5×5×64 | - | - | - | 1600×1 |
| Fully Connected 1 | 1600×1 | - | - | - | 100×1 |
| Fully Connected 2 | 100×1 | - | - | - | N×1 |

### 4.3.2 Agent design

In this layer, the state space $\mathcal{S}_1$, the action space $\mathcal{A}_1$, and the reward space $\mathcal{R}_1$ are defined respectively as follows.

**State space**: To describe the environment precisely and sufficiently, the environment state space $\mathcal{S}_1$ is represented by four state matrixes, namely, CAV locations, CAV speed, time-to-join (TTJ), and a map of the target lane, in this layer. Figure 8(a) shows an example of a formation zone where the red line on the rightmost denotes the stop line, and the black lines grid the interface into squares with side length of each square equal to lane width. Figure 8(b) illustrates that the target lane matrix is derived from the grids built upon the traffic environment. More details about transferring traffic environment into state matrixes can be found in our previous work.

**Action space**: The action space $\mathcal{A}_1 = \{1, 2, \dots, N\}$ is a set of the feasible sizes (i.e., the number of CAVs) of the platoon in the target lane, where $N$ is the largest feasible size corresponding to the largest feasible platoon length $L_n$. To ensure that the platoon length is smaller than the length of formation zone, $N$ shall meet the following constraints:

$$N = \arg\max_{n} L_n := \{n \mid L_n \leq L\} \tag{6}$$

$$L_n = \sum_{i=1}^{n} l_c^i + (n-1)d_h \tag{7}$$

where $l_c^i$ indicates the length of $i$-th CAV and $d_h$ denotes distance headway.

**Reward space**: In this layer, the optimization of three objectives, namely, efficiency, fairness and energy saving, is achieved by adopting the reward $\mathcal{R}_1$, which is defined as:

$$\mathcal{R}_1 = \frac{w_1 \cdot \sum_{i=0}^{m} PW_i{'} + w_2 \cdot \sum_{i=0}^{m} D_i{'} + w_3 \cdot \sum_{i=0}^{m} F_i{'}}{m} \tag{8}$$

where $m$ is the total number of CAVs during an action, and $w_1$, $w_2$ and $w_3$ are the weight coefficients which determine the relative importance of the three evaluation factors. Herein, $X' = \frac{X - X_{min}}{X_{max} - X_{min}}$, where $X_{min}$ and $X_{max}$ are obtained during a pre-training experiment. At the end of the action, the three factors, $PW_i$, $D_i$, and $F_i$, of each CAV are calculated, which are respectively explained as follows:

(1) $PW_i$ indicates penalized waiting time of the $i$-th CAV, which is calculated as:

$$PW_i = \left(\frac{W_i}{T_m}\right)^2 \tag{9}$$

where $W_i$ is the waiting time of the $i$-th CAV before the stop line and $T_m$ is a threshold value. The value of $PW_i$ larger than 1 (i.e., $W_i > T_m$) means that the $i$-th CAV has waited for too long and should be released.

(2) $D_i$ refers to the delay of the $i$-th CAV, which is calculated as

$$D_i = 1 - \frac{\bar{v}_i}{v_{max}} \tag{10}$$

where $\bar{v}_i$ is the average speed of the $i$-th CAV during the action and $v_{max}$ is the upper speed limit in the intersection.

(3) $F_i$ denotes the fuel consumption of the $i$-th CAV during the last action, which is measured according to Handbook on Emission Factors for Road Transport (HBEFA) (Keller et al., 2017).

**Remark 1.** At each timestamp, the actions in each lane are simultaneously conducted in the traffic environment. Herein, although $\mathcal{R}_1$ is a global reward contributed by the actions in all the lanes and should be allocated proportionally to each action, the reward of each action can be represented by $\mathcal{R}_1$ since the contribution of each action is equal at a timestamp. Hence, the reward defined in this way will not affect agent decision and learning.

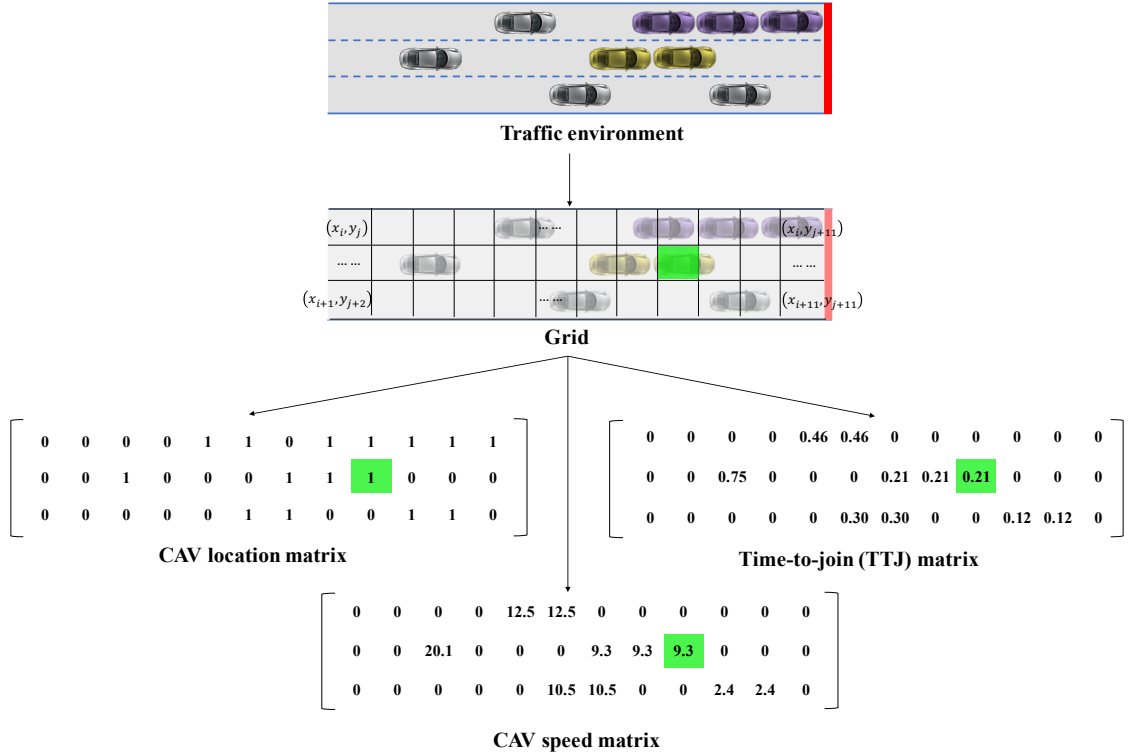

(a) CAV location matrix, CAV speed matrix and time-to-join (TTJ) matrix.

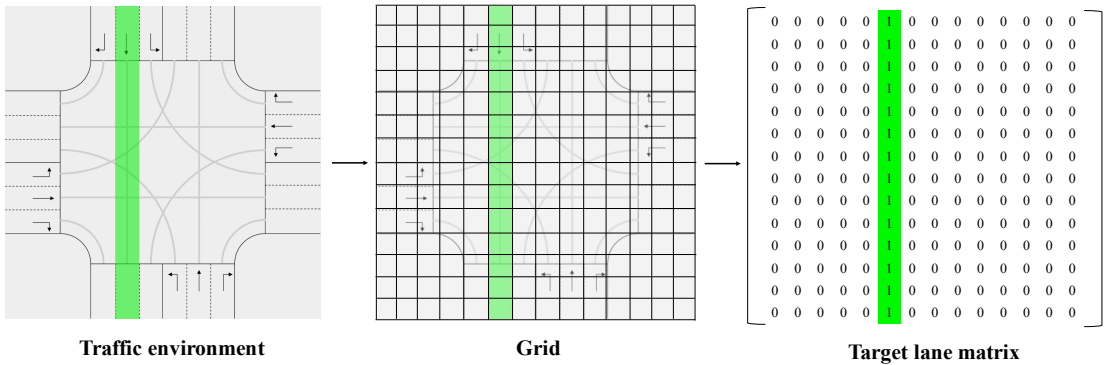

(b) Target lane matrix.

Figure 8. Example of transferring traffic environment into state matrixes.

## 4.4 Second layer: Platoon coordination layer

This layer addresses how to coordinate platoons to pass through the coordination zone. Each platoon travels led by its LVA, and each LVA is regarded as a DRL agent. Unlike the previous studies that solve this problem by controlling the acceleration of CAVs (as mentioned in Section 2), this study offers a simpler solution by controlling the passing priorities of platoons. In this way, the action space of the DRL agents can be greatly reduced, which enables the proposed model to be deployed in large-scale multi-agent systems (e.g., a busy intersection with multiple conflict platoons). Moreover, to reduce the complexity of multi-agent DRL problems, the model labels the platoons with "coordinated status" or "independent status" and processes them differently.

### 4.4.1 General process

As shown in Figure 9, the statuses of the platoons in the coordination zone are scanned and identified at each timestamp. The platoons labeled with independent status perform the independent action without considering coordination. The passing priorities of the platoons labeled with coordinated status are jointly determined while the coordination actions are separately performed. For example, Figure 10 illustrates the three scenarios that the platoon $P_1$ may meet, which are explained in Table 3. Herein, the desired grids indicate the desired location of a platoon at the next timestamp, while the overlaps of the desired grids of platoons are defined as conflict grids. In this layer, adopting DQN to determine the passing priorities is a key step. The DQN agent (i.e., the LVA involved in the coordination) generates four state matrixes by observing the coordination state. Using a CNN to process the state matrixes, the agent can derive an optimal coordination action based upon the passing priorities. The parameters of the CNNs in this layer with different granularities are presented in Table 4. Herein, the granularity has great effects on the representation of the platoons moving in the coordination zone. Since intersection involves two six-lane roads, three candidate values of granularity, namely, 6 ($6 \times 2^0$), 12 ($6 \times 2^1$), and 24 ($6 \times 2^2$), are considered in this study. Sensitivity analysis on model performances given the three candidate granularities is conducted in the experiments (see Section 5.2.1), upon which the optimal granularity is selected.

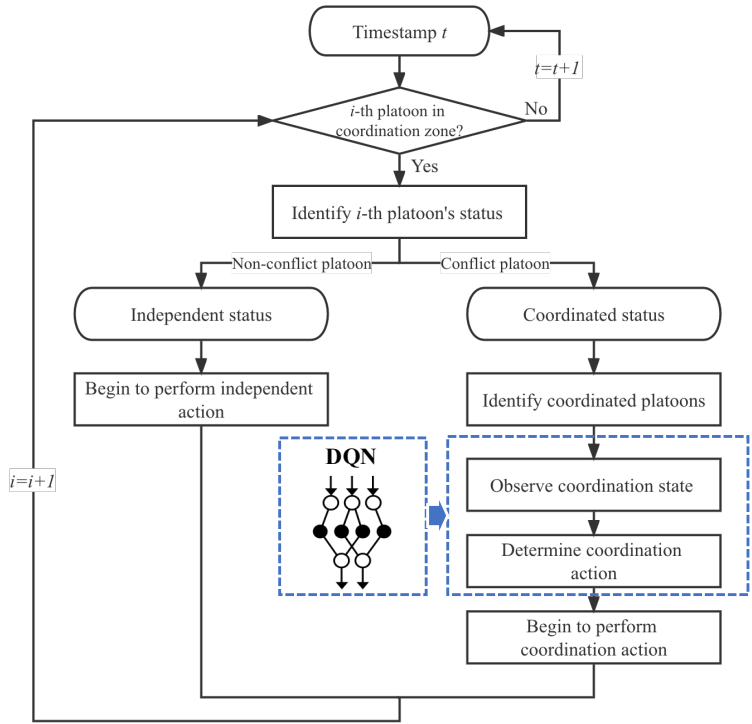

Figure 9. Flowchart of platoon coordination process.

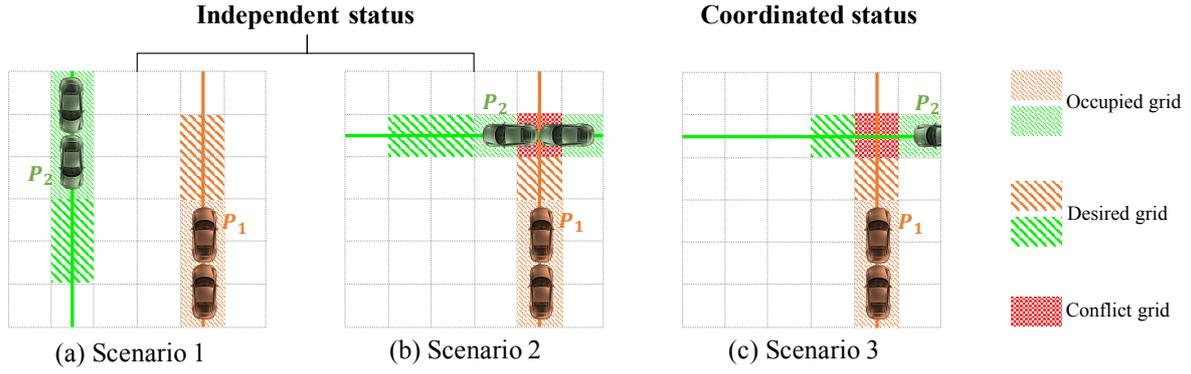

(a) Scenario 1  (b) Scenario 2  (c) Scenario 3

Figure 10. Illustrations of platoon statuses.

Table 3. Explanations of platoon statuses.

| Scenarios | Platoon status | Explanations (referring to Figure 10) |
|---|---|---|
| 1 | Independent status | The desired grids of $P_1$ are not overlapped with the desired grids of $P_2$. |
| 2 | Independent status | The desired grids of $P_1$ are being occupied by $P_2$. In this case, there is no need to coordinate since $P_1$ has to wait until $P_2$ exits the conflict grids. |
| 3 | Coordinated status | The desired grids of $P_1$ are overlapped with the desired grids of $P_2$. In this case, $P_1$ and $P_2$ shall coordinate to decide the passing priorities. |

Table 4. Parameters of CNNs in the second layer.

| Granularities | DQN layers | Input size | Kernel size | Strides | Filters | Output size |
|---|---|---|---|---|---|---|
| | Padding | 6×6×4 | - | - | - | 16×16×4 |
| | Convolution 1 | 16×16×4 | 3×3 | 1×1 | 32 | 14×14×32 |
| | Convolution 2 | 14×14×32 | 3×3 | 2×2 | 64 | 6×6×64 |
| $g = 6$ | Convolution 3 | 6×6×64 | 3×3 | 2×2 | 64 | 2×2×64 |
| | Reshape | 2×2×64 | - | - | - | 256×1 |
| | Fully Connected 1 | 256×1 | - | - | - | 16×1 |
| | Fully Connected 2 | 16×1 | - | - | - | k!×1 |
| | Padding | 12×12×4 | - | - | - | 20×20×4 |
| | Convolution 1 | 16×16×4 | 3×3 | 1×1 | 32 | 18×18×32 |
| | Convolution 2 | 18×18×32 | 3×3 | 3×3 | 64 | 6×6×64 |
| $g = 12$ | Convolution 3 | 6×6×64 | 3×3 | 2×2 | 64 | 2×2×64 |
| | Reshape | 2×2×64 | - | - | - | 256×1 |
| | Fully Connected 1 | 256×1 | - | - | - | 16×1 |
| | Fully Connected 2 | 16×1 | - | - | - | k!×1 |
| | Padding | 24×24×4 | - | - | - | 30×30×4 |
| | Convolution 1 | 30×30×4 | 3×3 | 1×1 | 32 | 28×28×32 |
| | Convolution 2 | 28×28×32 | 3×3 | 3×3 | 64 | 9×9×64 |
| | Convolution 3 | 9×9×64 | 3×3 | 2×2 | 64 | 4×4×64 |
| $g = 24$ | Max Pooling | 4×4×64 | - | 2×2 | - | 2×2×64 |
| | Reshape | 2×2×64 | - | - | - | 256×1 |
| | Fully Connected 1 | 256×1 | - | - | - | 16×1 |
| | Fully Connected 2 | 16×1 | - | - | - | k!×1 |

### 4.4.2 Agent design

In this layer, the state space $\mathcal{S}_2$, the action space $\mathcal{A}_2$, the reward space $\mathcal{R}_2$ and action-value update are respectively introduced as follows.

**State space**: Adopting the gridding method (refer to Definition 4 in Section 3.2) on coordination zone, the state space $\mathcal{S}_2$ of the agents can also be represented by four state matrixes. Figure 11 illustrates an example of transferring traffic environment into the state matrixes. In Figure 11, $P_1$ and $P_2$ are the two platoons with coordinated status, of which the state space is described by four matrixes, namely, current locations of conflict platoons, speed of conflict platoons, desired locations of conflict platoons, and current locations of other platoons. In the matrixes that represent current and desired locations of conflict platoons, the values of 1, 2, and 3 respectively denote the grid occupied by $P_1$, the grid occupied by $P_2$, and conflict grid. Also, $P_3$ is the platoon with independent status and is regarded as "other platoons" to $P_1$ and $P_2$, the current location of which is represented by the last matrix. In this study, the desired location $(x_{t+\Delta t}, y_{t+\Delta t})$ of an LVA at the timestamp $t + \Delta t$ is calculated as:

$$\begin{cases} x_{t+\Delta t} = x_t + v_t \sin \theta_t \Delta t + \frac{1}{2} a_t \sin \theta_t \Delta t^2 \\ y_{t+\Delta t} = y_t + v_t \cos \theta_t \Delta t + \frac{1}{2} a_t \cos \theta_t \Delta t^2 \end{cases} \quad (11)$$

where $(x_t, y_t)$ is the location of the LVA at the timestamp $t$, while $v_t$, $\theta_t$ and $a_t$ respectively denote the velocity, direction, and acceleration of the LVA, which are presented in Figure 12. It is assumed that the LVA keeps the same direction and acceleration during $\Delta t$. Then, the desired locations of the FVAs in the platoon can be obtained based on the desired location of its LVA.

**Action space**: The actions in responses to the three scenarios as mentioned in Table 3 and Figure 10 are listed in Table 5. There are $k!$ candidate actions, where $k$ is the number of conflict platoons involved in the coordination, in response to the coordinated status, upon which DRL is used to identify the optimal action.

Table 5. Action space with respect to different scenarios.

| Scenarios | Action types | Actions (referring to Figure 10) |
|---|---|---|
| 1 | Independent action | $P_1$ moves with maximum acceleration until achieving upper speed limit. |
| 2 | Independent action | $P_1$ Reduces speed with maximum deceleration. |
| 3 | Coordination action | The passing priorities of $P_1$ and $P_2$ are determined jointly, upon which the two platoons pass through the conflict grids in sequence. The platoon with higher passing priority moves with maximum acceleration, while the other stops and waits until the platoon with higher priority exits the conflict grids. |

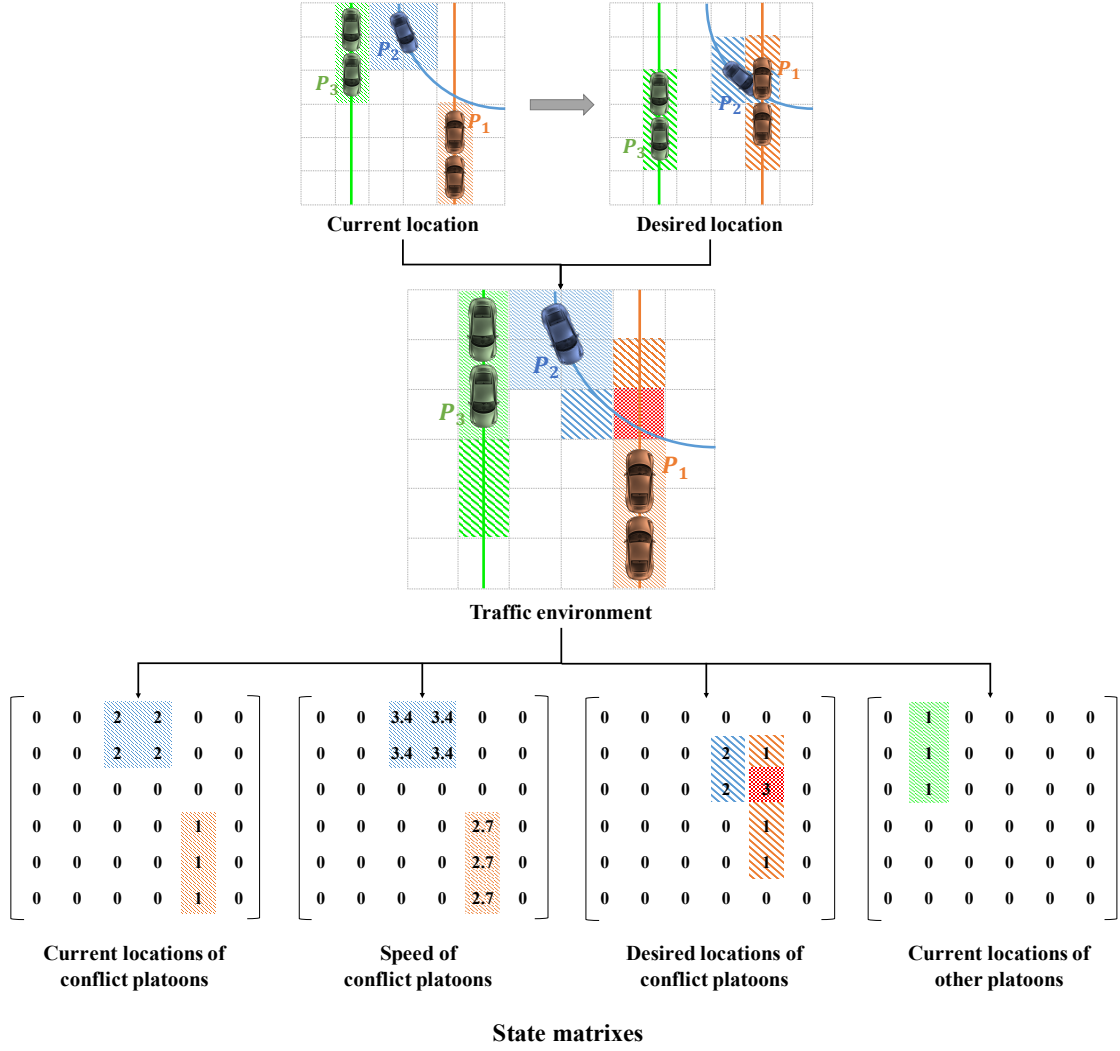

Figure 11. Example of transferring traffic environment into state matrixes.

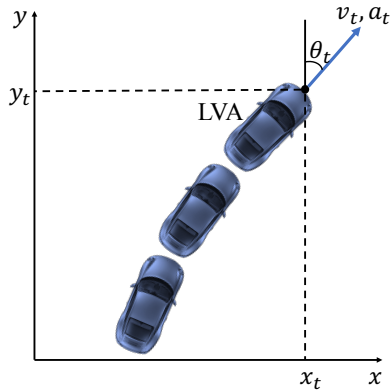

Figure 12. Illustration of current location, velocity, direction, and acceleration of an LVA.

**Reward space**: In this layer, the reward space $\mathcal{R}_2$ is obtained as:

$$\mathcal{R}_2 = -(CT + \frac{\sum_{i=0}^{n} PC_i}{k}) \tag{12}$$

where $k$ is the number of conflict platoons involved in the coordination. Herein, the two factors, namely, $CT$ and $PC_i$, are respectively introduced as follows:

(1) $CT$ is coordination time, which is defined as the time duration starting from the timestamp when MCD is reached (i.e., the coordinated status is identified) and ending at the timestamp when the last platoon in the coordination exits the conflict grids.

(2) $PC_i$ is the travel time of the *i*-th platoon, which is defined as the time duration starting from the timestamp when MCD is reached and ending at the timestamp when the *i*-th platoon exits the coordination zone.

**Remark 2.** As compared to $\mathcal{R}_1$ that takes multiple objectives (as mentioned in Section 4.3.2) into consideration, $\mathcal{R}_2$ considers only one objective, namely, time consumption, which is reflected by $CT$ and $PC_i$. This is because the durations of coordination processes are various but always short, which have extremely less impact on the objectives excluding time consumption.

**Action-value update:** When a coordination process is completed, the DRL agent will update the action-value function $Q$ to learn the optimal strategy. There are two ways to update $Q$, which are respectively adopted when a platoon meets the following two situations:

- Situation 1: The status of the platoon transfers from a coordinated status to another coordinated status. In this situation, $Q$ is updated by:

$$Q_{t+1}(s_t^C, a_t) = Q_t(s_t^C, a_t) + \alpha [\mathcal{R}_2 + \gamma \, max_{a_{t+1}} Q_t(s_{t+1}^C, a_{t+1}) - Q_t(s_t^C, a_t)] \quad (13)$$

where $s_t^C$ and $s_{t+1}^C$ respectively denote the states at the current timestamp and the next timestamp.

- Situation 2: The status of the platoon transfers from coordinated status to independent status. In this case, $Q$ is updated by:

$$Q_{t+1}(s_t^C, a_t) = \mathcal{R}_2 \quad (14)$$

**Remark 3.** As presented in Table 4, the platoon with independent status performs fixed independent action. Thus, action-value function update is not considered for the platoon with this status.

### 4.4.3 Deadlock: detection and punishment

Deadlock is a system failure that might be faced by multi-agent coordination. In this study, deadlock occurs when several platoons enter simultaneously into the coordination zone and the desired grids of each platoon are occupied by another platoon. However, such a failure has not been well considered and addressed in previous studies. In Figures 13(a) and 13(b), two examples of deadlocks respectively involving three and four platoons are illustrated. Using a directed graph to represent the relations among the coordinated platoons, deadlock can be detected if there is at least a cycle in the graph. Figures 13(c) and 13(d) respectively show the directed graphs corresponding to Figures 13(a) and 13(b), in which the cycles can be found.

Deadlock can lead to serious impact on traffic efficiency. Thus, it is necessary to train the model to learn how to avoid deadlocks. In this study, the approaches to detecting and addressing deadlocks are shown in Algorithm 2. Although deadlock might occur in the second layer of the COOR-PLT model, it is caused by the inappropriate platoon formation and releasing strategy implemented in the first layer. Hence, as presented at Line 6 in Algorithm 2, a punishment $R_{deadlock}$ is given to the DRL agent (i.e., IM) in order to reduce the potential of IM producing actions that may cause deadlock.

**Algorithm 2** Detection and solution of deadlock

1.    Represent relationship among platoons as a directed graph $G$
2.    **If** $G$ contains cycles **then**
3.       Deadlock is detected
4.       **For** each cycle $C_i$ **do**
5.          **For** each platoon $P_j$ in $C_i$ **do**
6.             Set $R_j = R_{deadlock}$ where $R_j$ is the reward in the first layer
7.             Store the experience $e_j = (s_j, a_j, r_j, s_{j+1})$ of $P_j$ in memory pool $D$
8.             Train the DRL agent in the first layer as Algorithm 1
9.             Remove $P_j$ from the coordination zone
10.          **End for**
11.       **End for**

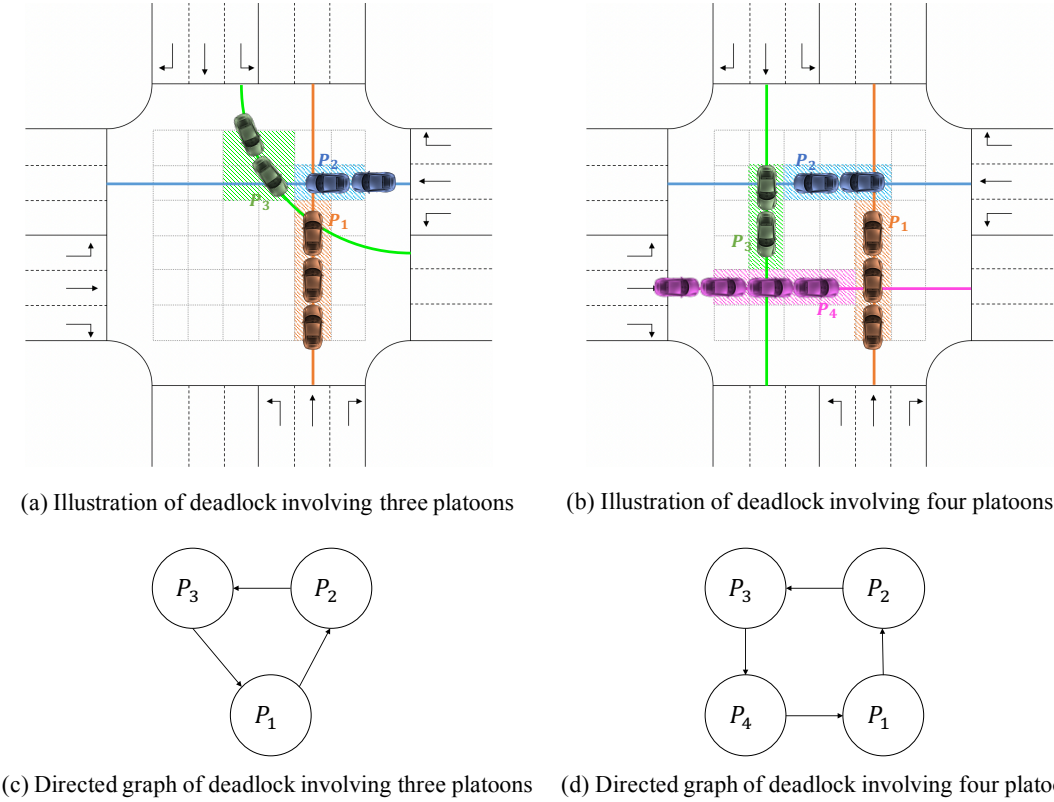

(a) Illustration of deadlock involving three platoons     (b) Illustration of deadlock involving four platoons

(c) Directed graph of deadlock involving three platoons     (d) Directed graph of deadlock involving four platoons

Figure 13. Examples of deadlocks.

# 5 Experiments

## 5.1 Experiment design and parameter setting

In this study, the proposed model is validated and examined on a traffic micro-simulator *Simulation of Urban Mobility* (SUMO). All the simulations are conducted using Python 3.6 on a computer with Core (TM) M-5Y71 CPU @ 1.20 GHz 1.40 GHz/RAM: 8.00 GB. A standard four-direction intersection with each direction consisting of a left-only lane, a straight-only lane, and a right-only lane is simulated as the environment. The parameters of the model and the simulated environment are listed in Table 6. It is assumed that the three factors (i.e., waited time, delay and fuel consumption) are equally important. Thus, the assigned value of the weight coefficients $w_1$, $w_2$ and $w_3$ are equal. As shown in Table 7, three traffic conditions at the intersection, namely, Conditions 1, 2, and 3, are designed and simulated. Conditions 1 and 2 respectively represent the traffic conditions with moderate and high traffic pressure, and Condition 3 indicates a traffic condition that mixes both traffic pressure. In real life, Condition 3 can represent the condition where there is sudden rise in traffic pressure which might be caused by special events (e.g., sports game, concert, etc.) and emergencies. It is worth noting that, in this study, the condition with low traffic pressure is not considered, since there are few conflicts that might occur between platoons at intersections, which has less demand for platoon coordination. To simulate the traffic environment more realistically, the flow rate in each lane is set differently, and the number of the CAVs that enter the simulated zone in a lane at a timestamp is assumed to follow Poisson distribution.

Table 6. Parameter settings.

| Categories | Parameters | Values | Explanations |
|---|---|---|---|
| Environment parameters | $l_c$ | 5.0 (m) | Length of CAVs |
| | $w_c$ | 1.8 (m) | Width of CAVs |
| | $l_{lane}$ | 2.5 (m) | Width of traffic lanes |
| | $S$ | 15.0 (m) | Length of the coordination zone |
| | $L$ | 200.0 (m) | Length of the formation zone |
| | $a_{max}$ | ±5 (m/s2) | Maximum acceleration and deceleration of CAVs |
| | $v_{max}$ | 20 (m/s) | Upper limit speed in the network |
| | $d_h$ | 1.0 m | Desired distance headway in the platoon |
| | $d_{\hat{h}}$ | 1.5 m | Minimum distance headway outside the platoon |
| Model parameters | T | 3600 (s) | Simulation duration in each episode |
| | M | 100 | Maximum number of training episodes |
| | $\alpha$ | 0.9 | Learning rate |
| | $\gamma$ | 0.9 | Discount factor |
| | $\varepsilon$ | 0.1 | $\varepsilon$-greedy strategy: 10% Exploration and 90% Exploitation |
| | N | 1000 | Maximum number of experiences in the memory pool |
| | batch_size | 32 | Size of memory extracted from the pool for learning each time |
| | O | 100 | Number of steps to observe before training process |
| | C | 200 | Frequency at which the parameters of the Target Net updates |
| | $T_m$ | 60 | Threshold value of $W_i$ |
| | $w_1$ | -1 | Weight coefficient of average penalized waiting time |
| | $w_2$ | -1 | Weight coefficient of average delay |
| | $w_3$ | -1 | Weight coefficient of average fuel consumption |
| | $R_{deadlock}$ | -10 | Punishment of deadlock |
| | $g$ | 6, 12, 24 | Granularity |

Table 7. Conditions of simulated traffic environment.

| Directions | Flow rate (veh/h) | |
| --- | --- | --- |
| | Moderate | High |
| North-South | 500 | 1000 |
| North-East | 400 | 800 |
| North-West | 300 | 600 |
| South-North | 450 | 900 |
| South-East | 600 | 1200 |
| South-West | 300 | 600 |
| East-North | 200 | 400 |
| East-South | 400 | 800 |
| East-West | 400 | 800 |
| West-North | 500 | 1000 |
| West-South | 200 | 400 |
| West-East | 300 | 600 |

| Conditions | Duration (s) | Flow rate |
| --- | --- | --- |
| 1 | 0-3600 | Moderate |
| 2 | 0-3600 | High |
| 3 | 0-1800 | Moderate |
| | 1800-3600 | High |

## 5.2 Experiment results

### 5.2.1 Sensitivity analysis of granularity

As mentioned in Section 4.4.1, granularity has great effects on the representation of the platoons moving in the coordination zone and the performance of the second layer. In this section, the sensitivity of the model to granularity is measured, upon which the granularity with which the model attains the best convergence performance is selected as the optimal granularity. Herein, the convergence performances of episode reward in the second layer, travel time, and occurrence frequency of deadlocks are respectively measured. There are three candidate values of granularity (i.e., 6, 12, and 24) as mentioned above. The convergence performances of the models given the three granularities in each traffic condition are presented in Figures 14, 15 and 16.

It is found that, in all the conditions, episode reward in the second layer shows a substantial increase while travel time decreases as granularity raises from 6 to 12. Moreover, deadlocks cannot be completely avoided during the first 100 episodes of training when granularity is 6, but being eliminated when granularity equals 12 and 24. Those findings can be explained from the following three perspectives: (1) higher granularity could help represent the locations of platoons more precisely, which facilitates the DRL agent to better understand the environment; (2) platoons occupy less grid space given higher granularity (see Figure 3), which can increase the throughput of intersection as more space is available for newly-entering platoons to occupy; and (3) for a platoon in coordination zone, the probability of the desired grids being occupied by another platoon is lower in high granularity, which can reduce the occurrence of deadlocks. There is no significant improvement in model performance when granularity raises from 12 to

24. However, higher granularity would increase computational expense. Therefore, to make a trade-off between precise representation of traffic environment and computational complexity, the granularity equal to 12 is selected for conducting the experiment.

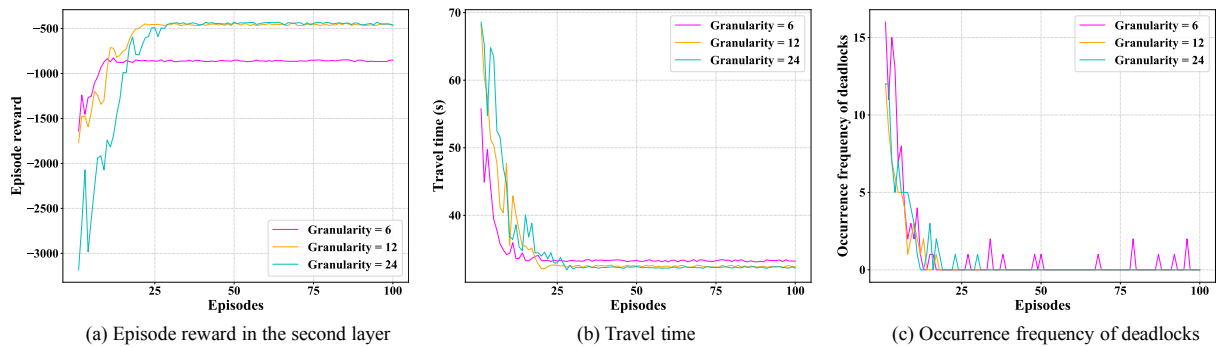

(a) Episode reward in the second layer  (b) Travel time  (c) Occurrence frequency of deadlocks

Figure 14. Convergence performances with respect to granularities in Condition 1.

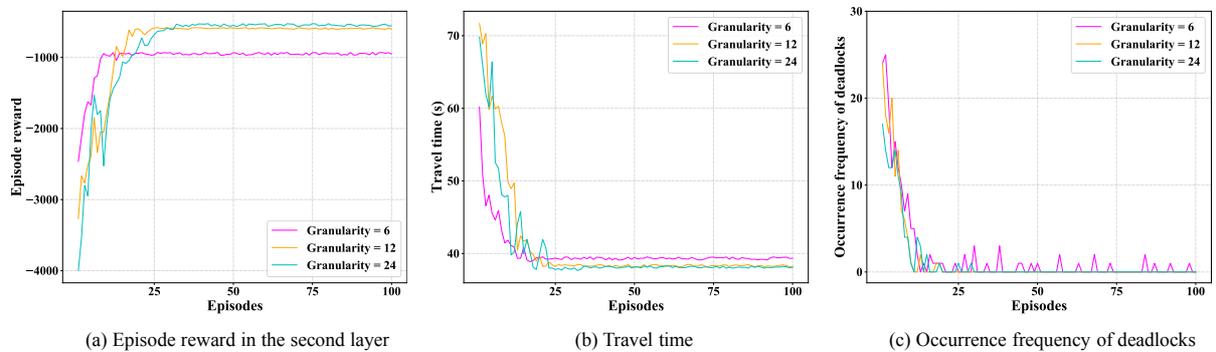

(a) Episode reward in the second layer  (b) Travel time  (c) Occurrence frequency of deadlocks

Figure 15. Convergence performances with respect to granularities in Condition 2.

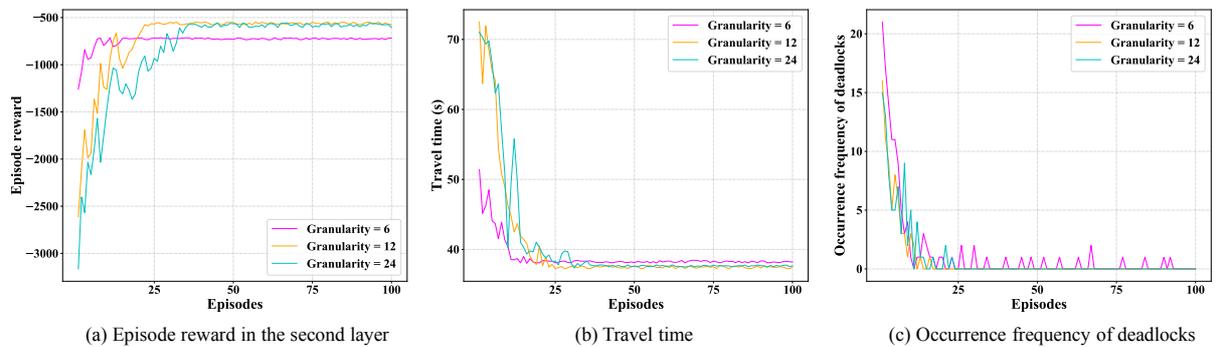

(a) Episode reward in the second layer  (b) Travel time  (c) Occurrence frequency of deadlocks

Figure 16. Convergence performances with respect to granularities in Condition 3.

### 5.2.2 Model performances

The model adopting the optimal granularity ($g = 12$) are evaluated from the following three perspectives:

- Model convergence

Figure 17 illustrates the convergence performances of the proposed model in the first 100 training episodes, in which five items are evaluated, namely, the episode rewards in the first and second layers, travel time, fuel consumption and occurrence of deadlocks. As presented in Figures 17(a) and 17(b), it is found that the episode rewards of the two layers of the proposed model converge within the first 25 episodes and perform steadily afterwards. Besides, as illustrated in Figures 17(c) to 17(e), the other three items also exhibit significant reduction with the increasing number of episodes. It is worth noting that there is no significant difference among the convergence speeds of the three traffic conditions for each item. As compared with the other two traffic conditions, the model of Condition 1 converges slightly faster because of the simpler settings of the condition. Therefore, it can be concluded that the proposed model is practicable as all the five evaluated items of the model converge at a satisfactory rate in different conditions. Further, the results of convergence performances demonstrate that the proposed model can fulfill the above-mentioned multi-objectives and achieve a trade-off among traffic efficiency, fairness and energy saving.

- Decision adaptability

Figure 18 shows the probability distribution of the platoon size in each traffic condition. It is found that the distributions of platoon size are different among the three conditions, and the platoon sizes are not evenly distributed in each condition, which manifests that the proposed model can adaptively decide the platoon size in response to the change of traffic conditions. Also, it is found that short platoons (e.g., the platoon with size equal to 3) are preferred under low traffic pressure (e.g., Condition 1), while longer platoons (e.g., the platoon with size equal to 7) are preferred under higher traffic pressure (e.g., Conditions 2 and 3). This finding can be explained by two of the objectives, namely, fairness and traffic efficiency, that the model aims to achieve during the training process. When traffic pressure is low, forming a long platoon might be time-consuming and reduce the traffic efficiency of the intersection. On the contrary, when traffic pressure is high, the time spent by forming a long platoon is reduced, upon which the traffic efficiency of the intersection might be improved. Moreover, it is worth noting that the probability distribution of Condition 3 can be seen as the result of fusing the distributions of Conditions 1 and 2, as Condition 3 mixes the first two traffic conditions.

- Deadlock avoidance

Figure 17(e) presents the convergence of deadlock occurrence in all the three traffic conditions, which demonstrates that the proposed model possesses the capability of avoiding deadlocks. In the three conditions, the occurrence frequency of deadlocks shows a more significant decline with the increasing number of episodes, which manifests the effectiveness and efficiency of the model learning to avoid deadlocks. For comparison, as presented in Figures 14(c), 15(c), and 16(c), deadlocks cannot be completely avoided when granularity equals 6, especially given the condition where the traffic pressure is higher. The incomplete avoidance of deadlock can be explained by that the deadlocks are more likely to be caused by long platoons, while longer platoons are preferred in the conditions with higher traffic pressure (e.g., Conditions 2 and 3) as mentioned above.

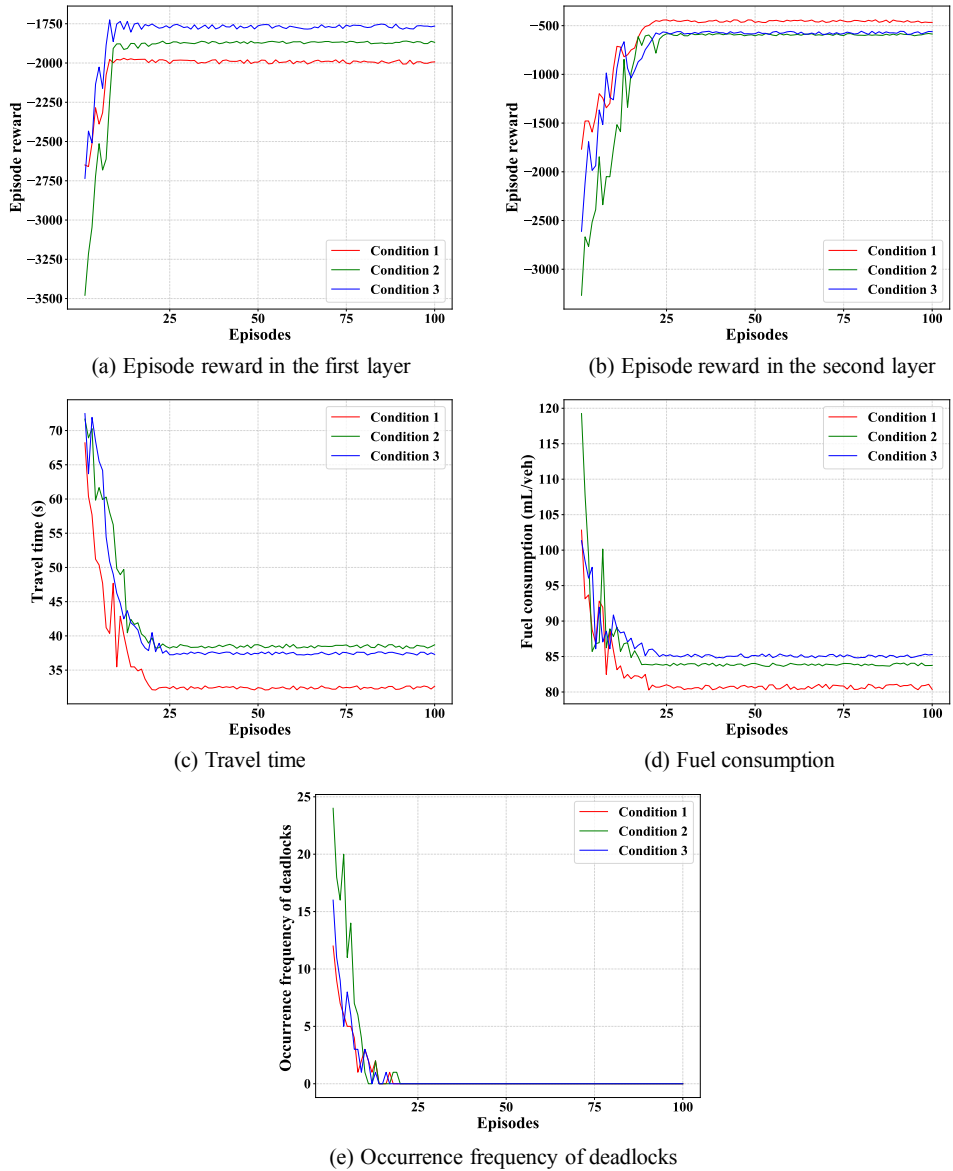

Figure 17. Convergence performance of training COOR-PLT.

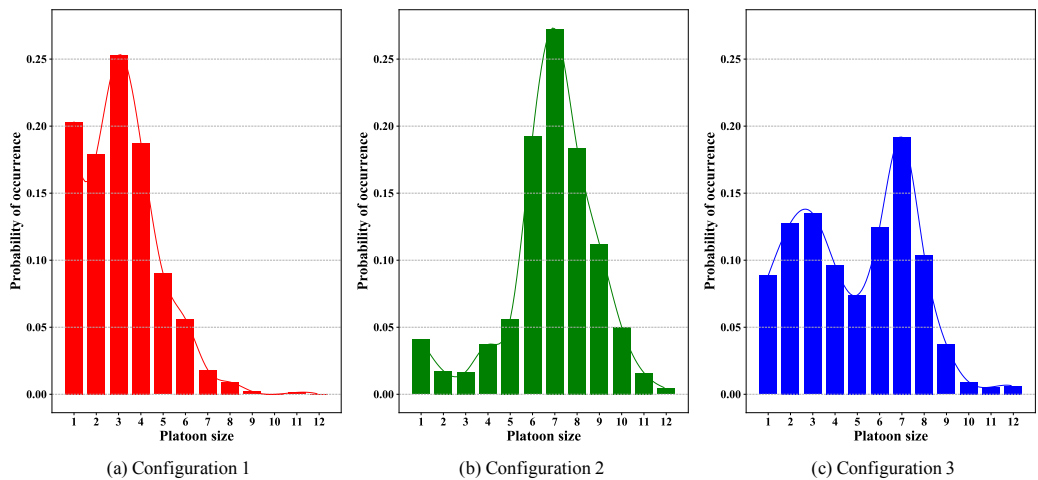

Figure 18. Probability distributions of platoon size.

## 5.3 Method comparisons

### 5.3.1 Baseline methods

There are in total six baseline methods employed for comparing with our proposed model. Two of them are the variants of our proposed model, which are used to verify the necessity of the techniques and strategies adopted in the proposed model. The other four are the existing state-of-the-art methods proposed in the previous studies, which are intended for demonstrating the superiority and advances of the proposed model. The two variants are respectively named as COOR-PLT-FP and COOR-PLT-RC. The difference between COOR-PLT-FP and the proposed model is that COOR-PLT-FP applies fixed platooning, upon which each platoon is formed in a fixed size equal to 3, instead of using DRL technique to determine the platoon size adaptively as the proposed model does in the first layer. The difference between COOR-PLT-RC and the proposed model is that COOR-PLT-RC uses random coordination to determine the passing priorities of conflict platoons rather than DRL technique in the second layer. The state-of-the-art methods are Webster method (Webster, 1958), Reservation-based AIM (Dresner and Stone, 2008), DCL-AIM (Wu et al., 2019), and INTEL-PLT (Li et al., 2022). Webster method and Reservation-based AIM are the two conventional methods. Webster method applies fixed signal timing control technique, while Reservation-based AIM employs FCFS policy for the reservations of individual CAVs. DCL-AIM and INTEL-PLT are the two RL-based methods. DCL-AIM adopts a coordination strategy for individual CAVs, while INTEL-PLT uses an adaptive CAV platooning method. Table 8 summarizes the similarities and differences between the baseline methods and the proposed model in details.

Table 8. Method comparisons.

| Categories | Methods | Control objects | Platooning strategy | Coordination strategy |
|---|---|---|---|---|
| Proposed model | COOR-PLT | CAV platoons | Adaptive platooning | DRL-based coordination |
| Variants | COOR-PLT-FP | CAV platoons | Fixed platooning | DRL-based coordination |
|  | COOR-PLT-RC | CAV platoons | Adaptive platooning | Random coordination |
| State-of-the-art methods | Webster method | Individual CAVs | - | - |
|  | Reservation-based AIM | Individual CAVs | - | - |
|  | DCL-AIM | Individual CAVs | - | RL-based coordination |
|  | INTEL-PLT | CAV platoons | Adaptive platooning | - |

### 5.3.2 Performance evaluation

The performances of the above methods on multiple aspects are evaluated for comparison. The variants are compared with the proposed model based on average travel time, fuel consumption and occurrence of deadlocks. The state-of-the-art methods are compared to the proposed model based on the above performance indicators excluding occurrence of deadlocks, since most of those methods do not involve deadlocks. It is worth mentioning that, for the methods in which coordination zone is gridded (i.e., Reservation-based AIM, DCL-AIM, the variants, and the proposed model), the value of granularity is set as 12 (i.e., $g = 12$). Besides, for the RL-based methods (i.e., DCL-AIM, INTEL-PLT, the variants, and the proposed model), the model after 100-episode training is used for evaluating the performances.

The performances of the variants are presented in Table 9. By comparing COOR-PLT-FP and COOR-PLT, it is found that involving adaptive platooning can sufficiently reduce travel time and fuel consumption in Conditions 2 and 3, but does not show much reduction in Condition 1. This is in line with the finding mentioned in Section 5.2.2 that larger platoon size is preferred in Conditions 2 and 3 while smaller platoon size (i.e., platoon size equal to 3) is preferred in Condition 1, which manifests that the fixed platooning as adopted by COOR-PLT-FP is suitable for limited traffic conditions. Moreover, it is found that both average travel time and fuel consumption of COOR-PLT are lower than those of COOR-PLT-RC in the three traffic conditions. The performance gaps of the above two factors between COOR-PLT and COOR-PLT-RC are wider in Conditions 2 and 3 than in Condition 1. This can be explained by that conflict movements become increasingly frequent with the rise of traffic pressure, and thus the superiority of the DRL-based coordination strategy over random coordination is more obvious. Further, as compared to the two variants, COOR-PLT demonstrates its superior capability of avoiding deadlocks. Hence, the comparisons between COOR-PLT and its variants manifest the necessities and benefits of adopting adaptive platooning and DRL-based coordination.

As presented in Table 10, the comparisons between COOR-PLT and the state-of-the-art methods manifest that COOR-PLT outperforms the state-of-the-art methods on the two aspects (i.e., travel time and fuel consumption) in the three conditions. The performances of both conventional methods (i.e., Webster method and reservation-based AIM) are unsatisfactory, which manifests that those methods with simple settings are unable to make efficient control in a CAV environment. The two RL-based methods (i.e., DCL-AIM and INTEL-PLT) outperform the conventional methods on the two aspects, which to an extent demonstrates the effectiveness of platooning and coordination strategies. Comparing DCL-AIM and INTEL-PLT, it is found that INTEL-PLT outperforms DCL-AIM on the two aspects in Condition 1 while underperforms in Conditions 2 and 3. It can be inferred that the performances of the two aspects depend more on coordination strategy than platooning strategy under low traffic pressure, while platooning strategy becomes more influential under higher traffic pressure. COOR-PLT outperforms the two RL-based methods in the three traffic conditions, as COOR-PLT integrates both platooning and coordination strategies and enhance their benefits by using DRL technique.

Table 9. Comparisons with variants of COOR-PLT.

| Methods | Condition 1 | | | Condition 2 | | | Condition 3 | | |
|---|---|---|---|---|---|---|---|---|---|
| | Average travel time (s) | Fuel consumption (mL/veh) | Occurrence frequency of deadlocks | Average travel time (s) | Fuel consumption (mL/veh) | Occurrence frequency of deadlocks | Average travel time (s) | Fuel consumption (mL/veh) | Occurrence frequency of deadlocks |
| COOR-PLT-FP | 33.97 | 81.04 | 9 | 59.73 | 90.52 | 19 | 53.51 | 92.79 | 23 |
| COOR-PLT-RC | 33.67 | 83.72 | 1 | 43.22 | 87.88 | 2 | 42.24 | 88.92 | 2 |
| COOR-PLT | 32.41 | 80.68 | 0 | 38.22 | 83.90 | 0 | 37.25 | 85.08 | 0 |

Table 10. Comparisons with state-of-the-art methods.

| Methods | Condition 1 | | Condition 2 | | Condition 3 | |
|---|---|---|---|---|---|---|
| | Average travel time (s) | Fuel consumption (mL/veh) | Average travel time (s) | Fuel consumption (mL/veh) | Average travel time (s) | Fuel consumption (mL/veh) |
| Webster method | 114.38 | 121.52 | 120.72 | 139.80 | 132.29 | 130.42 |
| Reservation-based AIM | 125.99 | 140.20 | 143.56 | 158.14 | 136.48 | 150.51 |
| DCL-AIM | 40.16 | 94.93 | 88.33 | 105.51 | 82.39 | 100.82 |
| INTEL-PLT | 58.28 | 99.56 | 79.25 | 100.89 | 72.16 | 101.34 |
| COOR-PLT | 32.41 | 80.68 | 38.22 | 83.90 | 37.25 | 85.08 |

# 6 Conclusion

In this study, we develop a DRL-powered hierarchical model for CAV control at signal-free intersections, named COOR-PLT. To the best of our knowledge, this study is one of the first attempts to integrate both adaptive platooning and coordination strategies to control CAVs at signal-free intersections. The model comprises two layers. In the first layer, a centralized control strategy is adopted to determine the optimal size for each platoon using global traffic information. In the second layer, a decentralized control strategy is proposed to coordinate the platoons passing through the intersection using local traffic information. The model comprehensively considers multiple objectives (i.e., efficiency, fairness and energy saving) when making decisions.

The experiments to validate and examine the proposed model are designed using SUMO. As a basis of the experiments, the optimal granularity is determined upon sensitivity analysis, which makes a trade-off between representation of traffic environment and computational complexity. The evaluation of model performances manifests that the proposed model: (1) has satisfactory convergence performances on episode rewards, travel time, fuel consumption, and occurrence of deadlocks; (2) has the ability to adaptively decide the platoon size in response to the change of traffic conditions; and (3) possesses the capability of completely avoiding deadlocks. Also, several methods are used for comparison with the proposed model. The comparison between the proposed model and its variants manifests the superiority of adopting adaptive platooning and DRL-based coordination in the model. The comparison between the proposed model and several state-of-the-art methods demonstrates the superior performances of the proposed model on reducing travel time and fuel consumption in different traffic conditions.

There are still several limitations of this study that could be resolved in the future. First, in this study, the proposed model is validated on a simulation platform, which has several differences with the practical application on the road. In the future, field experiments could be designed to further validate the model. Second, as a possible direction of future work, the advanced deep learning algorithms with better convergence performance and computational efficiency could be adopted to improve the model. Last but not the least, more traffic conditions could be taken into consideration, and the control strategies could be modified accordingly. For example, to make the CAV control strategies become more flexible, the CAVs could be allowed to join and leave a platoon when passing through the intersection zone.